\def\year{2019}\relax
\documentclass[letterpaper]{article} 
\usepackage{aaai19}  
\usepackage{times}  
\usepackage{helvet}  
\usepackage{courier}  
\usepackage{url}  
\usepackage{graphicx}  
\usepackage{amsmath}
\usepackage{amssymb}
\usepackage{adjustbox}
\usepackage{subfigure}
\usepackage{blindtext}
\usepackage{multirow}
\usepackage{bm}

\newcommand{\etal}{\text{et al.}}
\newcommand{\eg}{\text{e.g.,}}
\newcommand{\ie}{\text{i.e.,}}
\newcommand{\GAN}{GAttN}
\newcommand{\GRN}{GRN}

\begin{document}
%
\title{\GRN{}: Gated Relation Network to Enhance Convolutional Neural Network for Named Entity Recognition\thanks{Work was done during an internship at Microsoft Research.}}
\author{
Hui Chen$^{\dag}$, Zijia Lin$^\ddag$, Guiguang Ding$^{\dag}$, Jianguang Lou$^\ddag$, Yusen Zhang$^\S$, Borje Karlsson$^\ddag$ \\
$^\dag$Beijing National Research Center for Information Science and Technology(BNRist) \\
School of Software, Tsinghua University, Beijing, China \\
$^\S$School of Computer Science, Beijing Institute of Technology, Beijing, China\\
$^\ddag$Microsoft Research, Beijing, China \\
$\{$jichenhui2012, yusenzhang95$\}$@gmail.com, dinggg@tsinghua.edu.cn, $\{$zijlin, jlou, borje.karlsson$\}$@microsoft.com
}
\maketitle
\begin{abstract}
The dominant approaches for named entity recognition (NER) mostly adopt complex recurrent neural networks (RNN), \eg{} long-short-term-memory (LSTM). However, RNNs are limited by their recurrent nature in terms of computational efficiency. In contrast, convolutional neural networks (CNN) can fully exploit the GPU parallelism with their feed-forward architectures. However, little attention has been paid to performing NER with CNNs, mainly owing to their difficulties in capturing the long-term context information in a sequence. In this paper, we propose a simple but effective CNN-based network for NER, \ie{} gated relation network (\GRN{}), which is more capable than common CNNs in capturing long-term context. Specifically, in \GRN{} we firstly employ CNNs to explore the local context features of each word. Then we model the relations between words and use them as gates to fuse local context features into global ones for predicting labels. Without using recurrent layers that process a sentence in a sequential manner, our \GRN{} allows computations to be performed in parallel across the entire sentence. Experiments on two benchmark NER datasets (\ie{} CoNLL-2003 and Ontonotes 5.0) show that, our proposed \GRN{} can achieve state-of-the-art performance with or without external knowledge. It also enjoys lower time costs to train and test. We have made the code publicly available at https://github.com/HuiChen24/NER-GRN.
\end{abstract}

\section{Introduction}
Named Entity Recognition (NER) is one of the fundamental tasks in natural language processing (NLP). It is designed to locate a word or a phrase that references a specific entity, like person, organization, location, etc., within a text sentence. It plays a critical role in NLP systems for question answering, information retrieval, relation extraction, etc. And many efforts have been dedicated to the field.

Traditional NER systems mostly adopt machine learning models, such as Hidden Markov Model (HMM)~\cite{bikel1997nymble} and Conditional Random Field (CRF)~\cite{mccallum2003early}. Although these systems can achieve high performance, they heavily rely on hand-crafted features or task-specific resources~\cite{ma2016CNNBLSTMCRF}, which are expensive to obtain and hard to adapt to other domains or languages.

With the development of deep learning, recurrent neural network (RNN) along with its variants have brought great success to the NLP fields, including machine translation, syntactic parsing, relation extraction, etc. RNN has proven to be powerful in learning from basic components of text sentences, like words and characters~\cite{tran2017named}. Therefore, currently the vast majority of state-of-the-art NER systems are based on RNNs, especially long-short-term-memory (LSTM)~\cite{hochreiter1997long} and its variant Bi-directional LSTM (BiLSTM). For example, Huang \etal~\shortcite{huang2015bidirectional} firstly used a BiLSTM to enhance words' context information for NER and demonstrated its effectiveness. 

However, RNNs process the sentence in a sequential manner, because they typically factor the computation along the positions of the input sequence. As a result, the computation at the current time step is highly dependent on those at previous time steps. This inherently sequential nature of RNNs precludes them from fully exploiting the GPU parallelism on training examples, and thus can lead to higher time costs to train and test.

Unlike RNNs, convolutional neural network (CNN) can deal with all words in a feed-forward fashion, rather than composing representations incrementally over each word in a sentence. This property enables CNNs to well exploit the GPU parallelism. But in the NER community, little attention has been paid to performing NER with CNNs. It is mainly due to the fact that CNNs have the capacity of capturing local context information but they are not as powerful as LSTMs in capturing the long-term context information. Although the receptive field of CNNs can be expanded by stacking multiple convolution layers or using dilated convolution layers, the global context capturing issue still remains, especially for variant-sized text sentences, which hinders CNNs obtaining a comparable performance as LSTMs for NER.

In this paper, we propose a CNN-based network for NER, \ie{} Gated Relation Network (\GRN{}), which is more powerful than common CNNs for capturing long-term context information. Different from RNNs that capture the long-term dependencies in a recurrent component, our proposed \GRN{} aims to capture the dependencies within a sentence by modelling the relations between any two words. Modelling word relations permits \GRN{} to compose global context features without regard to the limited receptive fields of CNNs, enabling it to capture the global context information. This allows \GRN{} to reach comparable performances in NER versus LSTM-based models. Moreover, without any recurrent layers, \GRN{} can be trained by feeding all words concurrently into the neural network at one time, which can generally improve efficiency in training and test.

Specifically, the proposed \GRN{} is customized into 4 layers, \ie{} the representation layer, the context layer, the relation layer and the CRF layer. In the representation layer, like previous works, a word embedding vector and a character embedding vector extracted by a CNN are used as word features. In the context layer, CNNs with various kernel sizes are employed to transform the word features from the embedding space to the latent space. The various CNNs can capture the local context information at different scales for each word. Then, the relation layer is built on top of the context layer, which aims to compose a global context feature for a word via modelling its relations with all words in the sentence. Finally, we adopt a CRF layer as the loss function to train \GRN{} in an end-to-end manner.

To verify the effectiveness of the proposed \GRN{}, we conduct extensive experiments on two benchmark NER datasets, \ie{} CoNLL-2003 English NER and OntoNotes 5.0. Experimental results indicate that \GRN{} can achieve state-of-the-art performance on both CoNLL-2003 ($\rm F_1$=$91.44$ without external knowledge and $\rm F_1$=$92.34$ with ELMo \cite{peters2018deep} simply incorporated) and Ontonotes 5.0 ($\rm F_1$=$87.67$), meaning that using \GRN{}, CNN-based models can compete with LSTM-based ones for NER. Moreover, \GRN{} can also enjoy lower time costs for training and test, compared to the most basic LSTM-based model.

Our contributions are summarized as follows.
\begin{itemize}
\item We propose a CNN-based network, \ie{} gated relation network (\GRN{}) for NER. \GRN{} is a simple but effective CNN architecture with a more powerful capacity of capturing the global context information in a sequence than common CNNs.
\item We propose an effective approach for \GRN{} to model the relations between words, and then use them as gates to fuse local context features into global ones for incorporating long-term context information.
\item With extensive experiments, we demonstrate that the proposed CNN-based \GRN{} can achieve state-of-the-art NER performance comparable to LSTM-based models, while enjoying lower training and test time costs.

\end{itemize}

\section{Related Work}
Traditional NER systems mostly rely on hand-crafted features and task-specific knowledge. In recent years, deep neural networks have shown remarkable success in the NER task, as they are powerful in capturing the syntactic dependencies and semantic information for a sentence. They can also be trained in an end-to-end manner without involving subtle hand-crafted features, thus relieving the efforts of feature engineering.

\textbf{LSTM-based NER System.} Currently, most state-of-the-art NER systems employ LSTM to extract the context information for each word. Huang \etal~\shortcite{huang2015bidirectional} firstly proposed to apply a BiLSTM for NER and achieved a great success. Later~\citeauthor{ma2016CNNBLSTMCRF}~\shortcite{ma2016CNNBLSTMCRF} and \citeauthor{chiu2016named}~\shortcite{chiu2016named} introduced character-level representation to enhance the feature representation for each word and gained further performance improvement. MacKinlay \etal~\shortcite{tran2017named} proposed to stack BiLSTMs with residual connections between different layers of BiLSTM to integrate low-level and high-level features.~\citeauthor{Liu2018Empower}~\shortcite{Liu2018Empower} further proposed to enhance the NER model with a task-aware language model.

Though effective, the inherently recurrent nature of RNNs/LSTMs makes them hard to be trained with full parallelization. And thus here we propose a CNN-based network, \ie{} gated relation network (\GRN{}), to dispense with the recurrence issue. And we show that the proposed \GRN{} can obtain comparable performance as those state-of-the-art LSTM-based NER models while enjoying lower training and test time costs.

\textbf{Leveraging External Knowledge.} It has been shown that external knowledge can greatly benefit NER models. External knowledge can be obtained by means of external vocabulary resources or pretrained knowledge representation, etc. \citeauthor{chiu2016named}~\shortcite{chiu2016named} obtained $\rm F_1$=$91.62\%$ on CoNLL-2003 by integrating gazetteers. \citeauthor{peters2017semi}~\shortcite{peters2017semi} adopted a character-level language model pretrained on a large external corpus and gained substantial performance improvement. More recently, \citeauthor{peters2018deep}~\shortcite{peters2018deep} proposed ELMo, a deep language model trained with billions of words, to generate \textit{dynamic contextual} word features, and gained the latest state-of-the-art performance on CoNLL-2003 by incorporating it into a BiLSTM-based model. Our proposed \GRN{} can also incorporate external knowledge. Specifically, experiments show that, with ELMo incorporated, \GRN{} can obtain even slightly superior performance on the same dataset.

\textbf{Non-Recurrent Networks in NLP.} The efficiency issue of RNNs has started to attract attention from the NLP community. Several effective models have also been proposed to replace RNNs.~\citeauthor{gehring2017convolutional}~\shortcite{gehring2017convolutional} proposed a convolutional sequence-to-sequence model and achieved significant improvement in both performance and training speed.~\citeauthor{vaswani2017attention}~\shortcite{vaswani2017attention} used self-attention mechanism for machine translation and obtained remarkable translation performance. Our proposed \GRN{} is also a trial to investigate whether CNNs can get comparable NER performances as LSTM-based models with lower time costs for training and test. And different from \cite{gehring2017convolutional,vaswani2017attention}, we do not adopt the attention mechanism here, though \GRN{} is a general model and can be customized into the attention mechanism easily. 

Iterated dilated CNN (ID-CNN), proposed by \citeauthor{strubell2017fast}~\shortcite{strubell2017fast}, also aims to improve the parallelization of NER models by using CNNs, sharing similar ideas to ours. However, although ID-CNN uses dilated CNNs and stacks layers of them, its capacity of modelling the global context information for a variant-sized sentence is still limited, and thus its performance is substantially inferior to those of the state-of-the-art LSTM-based models. In contrast, our proposed \GRN{} can enhance the CNNs with much more capacity to capture global context information, which is mainly attributed to that the relation modelling approach in \GRN{} allows to model long-term dependencies between words without regard to the limited receptive fields of CNNs. And thus \GRN{} can achieve significantly superior performance than ID-CNN.

\section{Proposed Model}
In this section, we discuss the overall NER system utilizing the proposed \GRN{} in detail. To ease the explanation, we organize our system with 4 specific layers, \ie{} the representation layer, the context layer, the relation layer and the CRF layer. We will elaborate on these layers from bottom to top in the following subsections.


\subsection{Representation Layer}
Representation layer aims to provide informative features for the upper layers. The quality of features has great impacts on the system's performance. Traditionally, features are hand-crafted obeying some elaborative rules that may not be applicable to other domains. Therefore, currently many state-of-the-art approaches tend to employ deep neural networks for automatic feature engineering.

As previous works like \cite{Ye2018HSCRF}, the representation layer in \GRN{} is comprised of only word-level features and character-level features. In this paper, we use pre-trained \textit{static} word embeddings, \ie{} GloVe\footnote{http://nlp.stanford.edu/projects/glove/}~\cite{pennington2014glove}, as the initialized word-level feature. And during training, they will be fine-tuned. Here we denote the input sentence $\bm{s}$ as $\bm{s}=\{s_1,s_2,...,s_T\}$, where $s_i$ with $i=1,2,\ldots,T$ denotes the $i$th word in the sentence, and $T$ is the length of the sentence. We also use $\bm{y}=\{y_1,y_2,...,y_T\}$ to denote the corresponding entity labels for all words, \ie{} $y_i$ corresponding to $s_i$. With each word $s_i$ represented as a one-hot vector, its word-level feature $w_i$ is extracted as below:
\begin{equation}
\label{eq:word_embed}
w_i = E(s_i)
\end{equation}
where $E$ is the word embedding dictionary, initialized by the GloVe embeddings and fine-tuned during training.

\begin{figure}[!t] 
  \centering
  \includegraphics[width=0.85\linewidth]{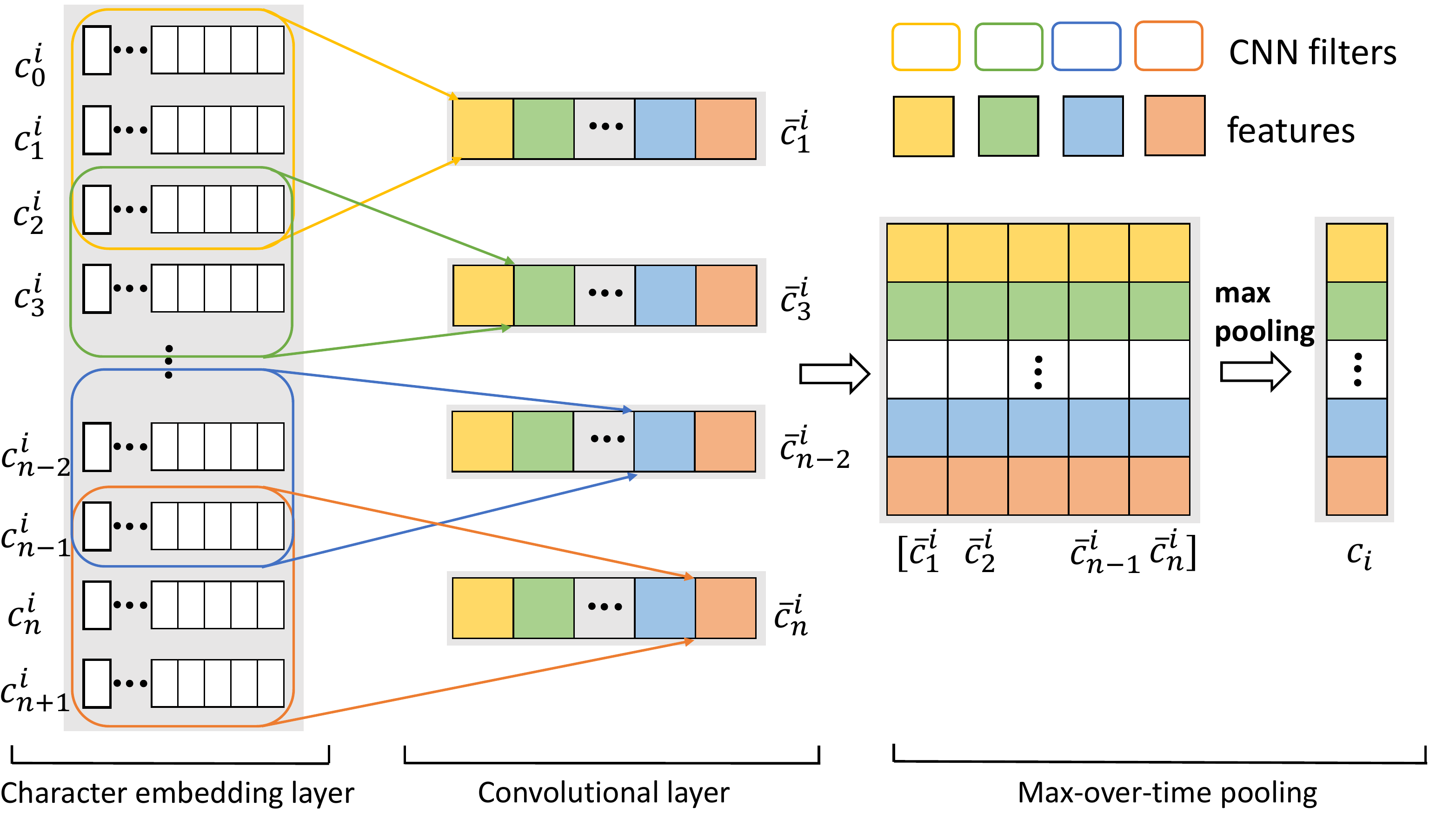}
  \caption{CNN to extract the character-level feature for a word. Best see in color.}
  \label{fig:char-cnn}
\end{figure}

Furthermore, we augment the word representation with the character-level feature, which can contribute to ease the out-of-vocabulary problem \cite{rei2016attending}. Same as \cite{ma2016CNNBLSTMCRF}, here we adopt a CNN to extract the character-level feature for each word $s_i$, as illustrated in Figure~\ref{fig:char-cnn}.

Specifically, the $j$-th character in the word $s_i$ containing $n$ characters is firstly represented as an embedding vector $c^i_j$ in a similar manner as Eq. \ref{eq:word_embed}, except that the character embedding dictionary is initialized randomly. Then we use a convolutional layer to involve the information of neighboring characters for each character, which is critical to exploiting n-gram features. Finally, we perform a max-over-time pooling operation to reduce the convolution results into a single embedding vector $c_i$:
\begin{equation}
\label{eq:char_cnn}
\begin{aligned}
&\bar{c}^i_j = \text{conv}([c^i_{j-k/2},...,c^i_j,...,c^i_{j+k/2}])\\
& c_i = \text{pooling}([\bar{c}^i_0,...,\bar{c}^i_j,...,\bar{c}^i_n])\\
\end{aligned}
\end{equation}
where $k$ is the kernel size of the convolutional layer. Here we fix $k=3$ as \cite{Ye2018HSCRF}.

Note that RNNs, especially LSTMs/BiLSTMs are also suitable to model the character-level feature. However, as revealed in \cite{yang2018design}, CNNs are as powerful as RNNs in modelling the character-level feature. Besides, CNNs can probably enjoy higher training and test speed than RNNs. Therefore, in this paper we just adopt a CNN to model the character-level feature.

We regard $c_i$ as the character-level feature for the word $s_i$. then we concatenate it to the word-level feature $w_i$ to derive the final word feature $z_i=[c_i,w_i]$.

\subsection{Context Layer}
Context layer aims to model the local context information among neighboring words for each word. The local context is critical for predicting labels, regarding that there could exist strong dependencies among neighboring words in a sentence. For example, a location word often co-occurs with prepositions like \textit{in}, \textit{on}, \textit{at}. Therefore, it is of necessity to capture the local context information for each word.

\begin{figure}[!t] 
  \centering
  \includegraphics[width=0.9\linewidth]{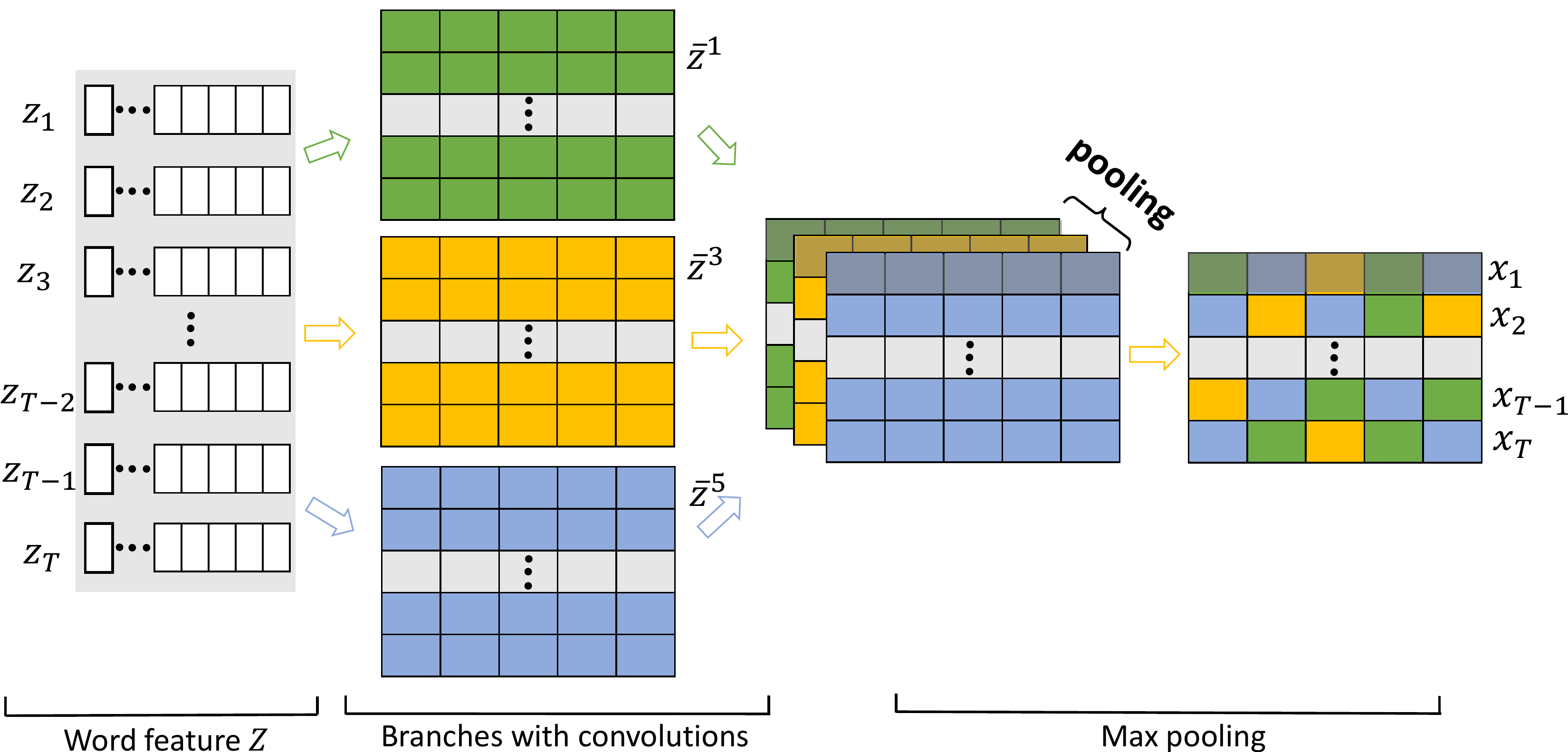}
  \caption{Branches with various convolutions for extracting the local context feature for words. Best see in colors.}
  \label{fig:inception}
\end{figure}

And it is obvious that the local dependencies are not limited within a certain distance. Therefore, we should enable the context layer to be adaptive to different scales of local information. Here, like InceptionNet~\cite{szegedy2015going}, we design the context layer with different branches, each being comprised of one certain convolutionaly layer. Figure~\ref{fig:inception} shows the computational process of the context layer.

Specifically, we use three convolutional layers with the kernel size being 1, 3, 5, respectively. After obtaining the word feature $Z=\{z_1,z_2,...,z_T\}$ of a sentence $\bm{s}$, each branch firstly extracts the local information $\bar{z}_i^k$ within a window-size $k$ for each word $s_i$. Then a max-pooling operation is employed to select the strongest channel-wise signals from all branches. To add the non-linear characteristic, we also apply \textit{tanh} after each branch.

\begin{equation}
\label{eq:inception}
\begin{aligned}
&\bar{z}_i^k = \text{conv}_k([z_{i-k/2},...,z_i,...,c_{i+k/2}])\\
& x_i = \text{pooling}([tanh(\bar{z}_i^1),tanh(\bar{z}_i^3),tanh(\bar{z}^5_i)])\\
\end{aligned}
\end{equation}
where $k \in \{1,3,5\}$ is the kernel size. For each $k$, we use $k/2$ zero-paddings to ensure that each word can get the corresponding context feature. Here, we consider the output $x_i$ of the context layer as the context feature for word $s_i$.

Although with various kernel sizes, the context layer can capture different kinds of local context information, it still struggles to capture the global one. However, we will show that with the gated relation layer described in the following subsection, the global context information can be realized by a fusion of the local one, thus tackling the shortcoming of the context layer.

\subsection{Relation Layer}
It has been shown that both short-term and long-term context information in a sequence is very critical in sequence learning tasks. LSTMs leverage the memory and the gating mechanism~\cite{hochreiter1997long} to capture both context information and gain significant success. However, conventional CNNs cannot well capture the long-term context information owing to the limited receptive fields, and thus existing CNN-based NER models cannot achieve comparable performance as LSTM-based ones.

\begin{figure}[!t] 
  \centering
  \includegraphics[width=0.85\linewidth]{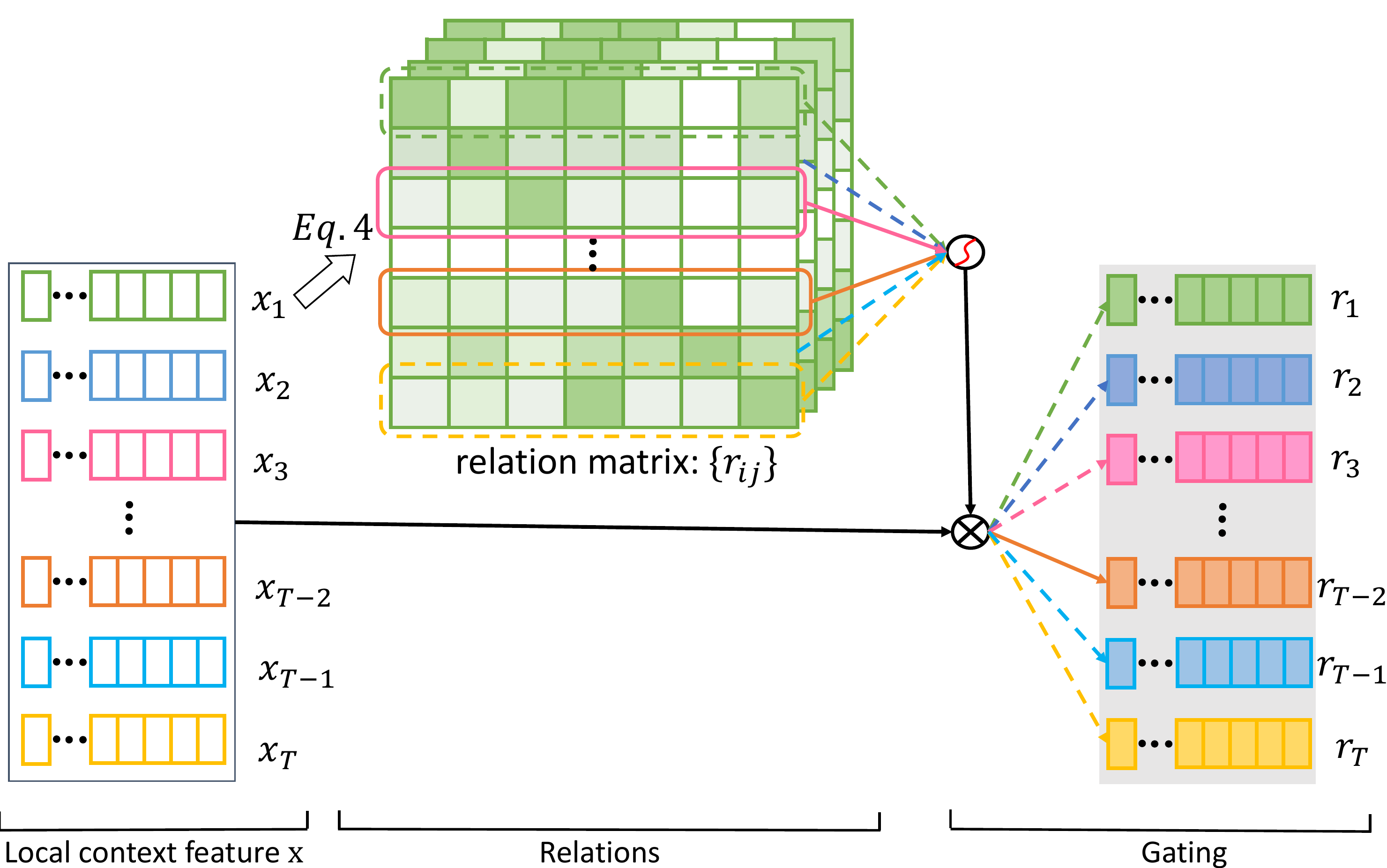}
  \caption{Gated relation layer in \GRN{} for composing the global context feature for each word. $r_{ij}$ denotes the relation score vector between word $s_i$ and word $s_j$. Best see in color.}
  \label{fig:relation}
\end{figure}

In this subsection, we introduce the gated relation layer in our proposed \GRN{}, which aims to enhance the conventional CNNs with global context information. Specifically, it models the relations between any two words in the sentence. Then, with the gating mechanism, it composes a global context feature vector by weighted-summing up the relation scores with their corresponding local context feature vectors, as shown in Figure~\ref{fig:relation}. Similar to the attention mechanism, our proposed \GRN{} is effective in modelling long-term dependencies without regard to the limited CNN receptive fields. And importantly, \GRN{} can allow computations to be performed in parallel across the entire sentence, which can generally help to reduce the time costs for training and test.



Given the local context features $\bm{x}=\{x_1,x_2,...,x_T\}$ from the context layer for a sentence $\bm{s}$, the relation layer firstly computes the relation score vector $r_{ij}$ between any two words $s_i$ and $s_j$, which is of the same dimension as any $x_i$. Specifically, it firstly concatenates the corresponding context features $x_i$ and $x_j$, and then uses a linear function with the weight matrix $W_{rx}$ and the bias vector $b_{rx}$ to obtain $r_{ij}$:
\begin{equation}
r_{ij} = W_{rx}[x_i;x_j] + b_{rx}
\end{equation}

Like~\cite{santoro2017simple}, we can directly average these relation score vectors as follows:
\begin{equation}
r_i = \frac{1}{T}\sum_{j=1}^T r_{ij}
\label{eq:relation_sum_up}
\end{equation}
where $r_i$ is the fused global context feature vector for the word $s_i$ by the direct feature fusion operation, \ie{} averaging in Eq. \ref{eq:relation_sum_up}. However, considering that non-entity words generally take up the majority of a sentence,  this operation may introduce much noise and mislead the label prediction. To tackle that, we further introduce the gating mechanism, and enable the relation layer to learn to select other dependent words adaptively. 
Specifically, for the word $s_i$, we firstly normalize all its relation score vectors $r_{ij}$ with a sigmoid function to reduce their biases. Then we sum up the normalized relation score vectors $r_{ij}$ with the corresponding local context feature vector $x_j \in \bm{x}=\{x_1,x_2,...,x_T\}$ of any other word $s_j$. And similar to Eq. \ref{eq:relation_sum_up}, finally we normalize the sum by the length of the sentence, \ie{} $T$.
\begin{equation}
r_i = \frac{1}{T}\sum_{j=1}^T \sigma(r_{ij}) \odot x_j
\label{eq:relation_gate}
\end{equation}
where $\sigma$ is a gate using sigmoid function, and $\odot$ means element-wise multiplication. Note that $r_{ij}$ is asymmetrical and different from $r_{ji}$, and the relation vector w.r.t $s_i$ itself, \ie{} $r_{ii}$, is also incorporated in the equation above. Therefore, with $r_i$ consisting of all the information of other words in the sentence, it can be seen as the global context feature vector for $s_i$.

In a way, \GRN{} can be seen as a channel-wise attention mechanism~\cite{chen2017sca}. However, instead of using a softmax function, we leverage the gating mechanism on the relation score vectors to decide how all the words play a part in predicting the label for the word $s_i$. We can also customize Eq.~\ref{eq:relation_gate} to the formula of attention with gating mechanism, where a gate is used to compute the attention weight for a word:
\begin{equation}
\begin{aligned}
&\alpha_{ij} = \sigma(W_x[x_i;x_j]+b_x)\\
&r_i = \frac{1}{T}\sum_{j=1}^T \alpha_{ij} *x_j\\
\end{aligned}
\label{eq:relation_attention}
\end{equation}
where $\alpha_{ij} \in R^1$ is an attention weight rather than a vector. 

To distinguish from the proposed \GRN{} (\ie{} Eq.~\ref{eq:relation_gate}), we name Eq.~\ref{eq:relation_sum_up} as Direct Fusion Network (DFN) and Eq.~\ref{eq:relation_attention} as Gated Attention Network (\GAN{}). We will consider DFN and \GAN{} as two of our baseline models to show the superiority of the proposed \GRN{}.

Here we also add a non-linear function for $r_i$ as follows.
\begin{equation}
\label{eq:pi}
p_i = tanh(r_i)
\end{equation}
And we define $p_i$ as the final predicting feature for word $s_i$.

\subsection{CRF Layer}
Modelling label dependencies is crucial for NER task~\cite{ma2016CNNBLSTMCRF,Liu2018Empower}. Following \cite{ma2016CNNBLSTMCRF,huang2015bidirectional}, we employ a conditional random field (CRF) layer to model the label dependencies and calculate the loss for training \GRN{}.

Formally, for a given sentence $\bm{s}=\{s_1,s_2,...,s_T\}$ and its generic sequence of labels $\bm{y}=\{y_1,y_2,...,y_T\}$, we firstly use $\mathcal{Y}(\bm{s})$ to denote the set of all possible label sequences for $\bm{s}$. The CRF model defines a family of conditional probability $p(\bm{y}|\bm{s})$ over all possible label sequences $\bm{y}$ given $\bm{s}$:
\begin{equation}
p(\bm{y}|\bm{s})=\frac{\prod_{i=1}^T\phi_i(y_{i-1},y_i,\bm{s})}{\sum_{y'\in \mathcal{Y}(\bm{s})}\prod_{i=1}^T\phi_i(y'_{i-1},y'_i,\bm{s})}
\end{equation}
where $\phi_i(y_{i-1},y_i,s)=\text{exp}(f(s_i, y', y))$ with $f$ being a function that maps words into labels:
\begin{equation}
f(s_i, y', y) = W_{y}p_i+b_{y',y}
\end{equation}
where $p_i$ is derived as Eq.~\ref{eq:pi}, $W_{y}$ is the predicting weights w.r.t $y$ and $b_{y',y}$ is the transition weight from $y'$ to $y$. Both $W_{y}$ and $b_{y',y}$ are parameters to be learned.

Loss of the CRF layer is formulated as follows.
\begin{equation}
L=-\sum_{\bm{s}} \log p(\bm{y}|\bm{s})
\end{equation}

And for decoding, we aim to find the label sequence $\bm{y}*$ with the highest conditional probability:
\begin{equation}
\bm{y}*=\mathrm{\arg\max}_{\bm{y}\in \mathcal{Y}(\bm{s})}p(\bm{y}|\bm{s})
\end{equation}
which can be efficiently derived via Viterbi decoding.

\section{Experiments}
To verify the effectiveness of the proposed \GRN{}, we conduct extensive experiments on two benchmark NER datasets: CoNLL-2003 English NER \cite{tjong2003introduction} and OntoNotes 5.0 \cite{hovy2006ontonotes,pradhan2013towards}.
\begin{itemize}
\item \textbf{CoNLL-2003 English NER} consists of 22,137 sentences totally and is split into 14,987, 3,466 and 3,684 sentences for the training set, the development set and the test set, respectively. It includes annotations for 4 types of named entities: \texttt{PERSON}, \texttt{LOCATION}, \texttt{ORGANIZATION} and \texttt{MISC}. 
\item \textbf{OntoNotes 5.0} consists of much more (76,714) sentences  from a wide variety of sources (telephone conversation, newswire, etc.). Following~\cite{chiu2016named}, we use the portion of the dataset with gold-standard named entity annotations, and thus excluded the New Testaments portion. It also contains much more kinds of entities, including \texttt{CARDINAL}, \texttt{MONEY}, \texttt{LOC}, \texttt{PRODUCT}, etc.
\end{itemize}
Table~\ref{tab:dataset} shows some statistics of both datasets. Following~\cite{ma2016CNNBLSTMCRF}, we use the BIOES sequence labelling scheme instead of BIO for both datasets to train models. As for test, we convert the prediction results back to the BIO scheme and use the standard CoNLL-2003 evaluation script to measure the NER performance, \ie{} $\rm F_1$ scores, etc.

\begin{table}[t!]
  \centering
    \begin{adjustbox}{max width=0.8\columnwidth}
    \scalebox{0.9}{
    \begin{tabular}{|l|c|c|c|c|}
      \hline
       dataset & &Train &Dev &Test  \\ \hline
      \multirow{3}{*}{CoNLL-2003} &Sentence &14,987 &3,466 &3,684 \\
      &Token &204,567 &51,578 &46,666 \\
      &Entity &23,499 &5,942 &5,648 \\
      \hline
      \multirow{3}{*}{OntoNotes 5.0} &Sentence &59,924 &8,528 &8,262 \\
      &Token &1,088,503 &147,724 &152,728 \\
      &Entity &81,828 &11,066 &11,257 \\
      \hline
    \end{tabular}}
    \end{adjustbox}
    \caption{Statistics of CoNLL-2003 and Ontonotes 5.0.}
    \label{tab:dataset}
\end{table}

\subsection{Network Training}
We implement our proposed \GRN{} with the Pytorch library~\cite{paszke2017automatic}. And we set the parameters below following \cite{ma2016CNNBLSTMCRF}.

\textbf{Word Embeddings.} The dimension of word embedding is set as $100$. And as mentioned, we initialize it with Stanford's publicly available GloVe 100-dimensional embeddings. We include all words of GloVe when building the vocabulary, besides those words appearing at least $3$ times in the training set. For words out of the vocabulary (denoted as \texttt{UNK}) or those not in GloVe, we initialize their embeddings with kaiming uniform initialization \cite{he2015delving}. 

\textbf{Character Embeddings.} We set the dimension of character embeddings as $30$, and also initialize them with kaiming uniform initialization.

\textbf{Weight Matrices and Bias Vectors.} All weight matrices in linear functions and CNNs are initialized with kaiming uniform initialization, while bias vectors are initialized as $0$.

\textbf{Optimization.} We employ mini-batch stochastic gradient descent with momentum to train the model. The batch size is set as $10$. The momentum is set as $0.9$ and the initial learning rate is set as $0.02$. We use learning rate decay strategy to update the learning rate during training. Namely, we update the learning rate as $\frac{0.02}{1+\rho*t}$ at the $t$-th epoch with $\rho=0.02$. We train each model on training sets with $200$ epochs totally, using dropout = $0.5$. For evaluation, we select its best version with the highest performance on the development set and report the corresponding performance on the test set. To reduce the model bias, we carry out 5 runs for each model and report the average performance and the standard deviation.

\textbf{Network Structure.} The output channel number of the CNN in Eq.~\ref{eq:char_cnn} and Eq.~\ref{eq:inception} is set as 30 and 400, respectively.

\begin{table}[t!]
  \centering
    \begin{adjustbox}{max width=\columnwidth}
    \begin{tabular}{|l|c|c|c|}
      \hline
       Model & Mean($\pm std$) $\rm{F_1}$ &Max $\rm{F_1}$ & Mean P/R \\ \hline
       \cite{collobert2011natural} & 88.67 & &\\ \hline
       \cite{luo2015joint} & 89.90 & &\\ \hline
       \cite{chiu2016named} & 90.91 $\pm$ 0.20 & &90.75 / 91.08 \\ \hline
       \cite{zhuo2016segment} & 88.12  & &\\ \hline
       \cite{rei2016attending} &84.09 & &\\ \hline
       \cite{lample2016neural} & 90.94 & &\\ \hline
       \cite{ma2016CNNBLSTMCRF} & 91.21 & &\textbf{91.35} / 91.06 \\ \hline
       \cite{rei2017semi} & 86.26 & &\\ \hline
       \cite{zukov2017neural} &89.83 & &\\ \hline
       \cite{liu2017capturing} &89.5 &  &\\ \hline
       \cite{peters2017semi} & 90.87 & &\\ \hline
       \cite{Liu2018Empower} & 91.24 $\pm$ 0.12 &91.35 &\\ \hline
       \cite{Ye2018HSCRF} &91.38 $\pm$ 0.10 &91.53 &\\ \hline
       ID-CNN~\cite{strubell2017fast} &90.54 $\pm$ 0.18 & &\\ \hline
       CNN-BiLSTM-CRF & 91.17 $\pm$ 0.18 &91.45 &\\ \hline
       CNN-BiLSTM-Att-CRF &90.24 $\pm$ 0.15 &90.48 &\\ \hline
       \textbf{\GRN{}} & \textbf{ 91.44 $\pm$ 0.16} &\textbf{91.67} & 91.31 / \textbf{91.57}\\ \hline
       \hline
       \cite{collobert2011natural}* & 89.59 & &\\ \hline
       \cite{luo2015joint}* & 91.2 &  &\\ \hline
       \cite{chiu2016named}* & 91.62 $\pm$ 0.33 & & 91.39 / 91.85 \\ \hline
       \cite{peters2017semi}* & 91.93 $\pm$ 0.19  & &\\ \hline
       (Tran et al. 2010)* & 91.66  & &\\ \hline
       (Yang et al. 2017)* & 91.26 &  &\\ \hline
       \cite{peters2018deep}* & 92.22 $\pm$ 0.10 & &\\ \hline
       \textbf{\GRN{}*} & \textbf{92.34 $\pm$ 0.10} &\textbf{92.45} &\textbf{92.04} / \textbf{92.65}\\ \hline
    \end{tabular}
    \end{adjustbox}
    \caption{Performance comparison on CoNLL-2003. * indicates models utilizing external knowledge beside the CoNLL-2003 training set and pre-trained word embeddings. P/R denotes precision and recall.}
    \label{tab:compare_conll}
\end{table}

\subsection{Performance Comparison}
Here we first focus on the NER performance comparison between the proposed \GRN{} and the existing state-of-the-art approaches.

\textbf{CoNLL-2003.} We compare \GRN{} with various state-of-the-art LSTM-based NER models, including~\cite{Liu2018Empower,Ye2018HSCRF}, etc. We also compare \GRN{} with ID-CNN~\cite{strubell2017fast}, which also adopts CNNs without recurrent layers for NER. Furthermore, considering that some state-of-the-art NER models exploit external knowledge to boost their performance, here we also report the performance of \GRN{} with ELMo~\cite{peters2018deep} incorporated as the external knowledge. Note that ELMo is trained on a large corpus of text data and can generate \textit{dynamic contextual} features for words in a sentence. Here we simply concatenate the output ELMo features to the word feature in \GRN{}. The experimental results are reported in Table~\ref{tab:compare_conll}, which also includes the max $\rm F_1$ scores, mean precision and recall values if available. Note that CNN-BiLSTM-CRF is our re-implementation of~\cite{ma2016CNNBLSTMCRF}, and we obtain comparable performance as that reported in the paper. Therefore, by default we directly compare \GRN{} with the reported performance of compared baselines. It should also be noticed that, since the relation layer in \GRN{} can be related to the attention mechanism, here we also include some attention-based baselines, \ie, \cite{rei2016attending} and \cite{zukov2017neural}, and we further introduce a new baseline termed CNN-BiLSTM-Att-CRF, which adds a self-attention layer for CNN-BiLSTM-CRF as ~\cite{zukov2017neural}.

As shown in Table~\ref{tab:compare_conll}, compared with those LSTM-based NER models, the proposed \GRN{} can obtain comparable or even slightly superior performance, with or without the external knowledge, which well demonstrates the effectiveness of \GRN{}. And compared with ID-CNN, our proposed \GRN{} can defeat it at a great margin in terms of $\rm{F_1}$ score. We also try to add ELMo to the latest state-of-the-art model of \cite{Ye2018HSCRF} based on their published codes, and we find that the corresponding $\rm{F_1}$ score is $91.79\pm0.08$, which is substantially lower than that of \GRN{}.


\textbf{OntoNotes 5.0.}
On OntoNotes 5.0, we compare the proposed \GRN{} with NER models that also reported performance on it,  including~\cite{chiu2016named,shen2017deep,durrett2014joint}, etc. As shown in Table~\ref{tab:compare_OntoNote}, \GRN{} can obtain the state-of-the-art NER performance on OntoNotes 5.0, which further demonstrates its effectiveness.

\begin{table}[t!]
  \centering
    \begin{adjustbox}{width=0.95\columnwidth}
    \begin{tabular}{|l|c|c|}
      \hline
       Model & Mean($\pm std$) $\rm{F_1}$ & Mean P/R  \\ \hline
       \cite{chiu2016named} & 86.28 $\pm$ 0.26 & 86.04 / 86.53\\ \hline
       \cite{shen2017deep} & 86.63 $\pm$ 0.49 & \\ \hline
       \cite{durrett2014joint} & 84.04 & 85.22 / 82.89\\ \hline
       \cite{passos2014lexicon} & 82.30 & \\ \hline
       \cite{ratinov2009design} & 83.45 & \\ \hline
       CNN-BiLSTM-Att-CRF &87.25$\pm$0.17 &\\ \hline
       ID-CNN~\cite{strubell2017fast} & 86.84 $\pm$ 0.19 &  \\ \hline
       \GRN{} & \textbf{87.67 $\pm$ 0.17} & \textbf{87.79} / \textbf{87.56}\\ \hline
    \end{tabular}
    \end{adjustbox}
    \caption{Performance comparison on OntoNotes 5.0. P/R denotes precision and recall.}
    \label{tab:compare_OntoNote}
\end{table}

\begin{table}[t!]
  \centering
    \begin{adjustbox}{max width=0.9\columnwidth}
    \begin{tabular}{|l|l|c|c|}
      \hline
       &Model & Mean($\pm std$) $\rm{F_1}$ & $\rm{F_1}$ Drop  \\ \hline
       \multirow{2}{*}{context} &\GRN{} w/o context & 88.36 $\pm$ 0.21 & 3.08 \\ \cline{2-4}
        &\GRN{} w/ $\rm branch_3$ & 90.88 $\pm$ 0.22 & 0.56 \\ \hline
       \hline
       \multirow{3}{*}{relation} &\GRN{} w/o relation & 90.13 $\pm$ 0.28 & 1.31\\ \cline{2-4}
       &DFN & 90.72 $\pm$ 0.06  & 0.72\\ \cline{2-4}
       &\GAN{} & 87.11 $\pm$ 0.25  & 4.33\\ \hline
       \hline
       Full&\GRN{} & \textbf{91.44 $\pm$ 0.16}  & 0\\ \hline
    \end{tabular}
    \end{adjustbox}
    \caption{Ablation study on CoNLL-2003.}
    \label{tab:compare_ablation_conll}
\end{table}

\begin{table}[t!]
  \centering
    \begin{adjustbox}{max width=0.9\columnwidth}
    \begin{tabular}{|l|l|c|c|}
      \hline
       &Model & Mean($\pm std$) $\rm{F_1}$ & $\rm{F_1}$ Drop  \\ \hline
       \multirow{2}{*}{context} &\GRN{} w/o context & 82.21 $\pm$ 0.23 & 5.46 \\ \cline{2-4}
        &\GRN{} w/ $\rm branch_3$ & 86.66 $\pm$ 0.21 & 1.01 \\ \hline
       \hline
       \multirow{3}{*}{relation} &\GRN{} w/o relation & 85.87 $\pm$ 0.16 & 1.8 \\ \cline{2-4}
       &DFN & 85.81 $\pm$ 0.14  & 1.86\\ \cline{2-4}
       &\GAN{} & 79.83 $\pm$ 0.37  & 7.83\\ \hline
       \hline
       Full&\GRN{} & \textbf{87.67 $\pm$ 0.17}  & 0\\ \hline
    \end{tabular}
    \end{adjustbox}
    \caption{Ablation study on OntoNotes 5.0.}
    \label{tab:compare_ablation_ontonote}
\end{table}

Overall, the comparison results on both CoNLL-2003 and OntoNotes 5.0 well indicate that our proposed \GRN{} can achieve state-of-the-art NER performance with or without external knowledge. It demonstrates that, using \GRN{}, CNN-based models can compete with LSTM-based ones for NER.

\subsection{Ablation Study}
Here we study the impact of each layer on \GRN{}. Firstly, we analyze the context layer by introducing two baseline models: (1) \GRN{} w/o context: wiping out the context layer and building the relation layer on top of the representation layer directly; (2) \GRN{} w/ $\rm branch_3$: removing branches in the context layer, except the one with kernel size $= 3$. Then to analyze the relation layer and the importance of gating mechanism in it, we compare \GRN{} with: (1) \GRN{} w/o relation: wiping out the relation layer and directly building the CRF layer on top of the context feature; (2) DFN (see Eq.~\ref{eq:relation_sum_up}); (3) \GAN{} (see Eq.~\ref{eq:relation_attention}). All compared baselines use the same experimental settings as \GRN{}. Table~\ref{tab:compare_ablation_conll} and Table~\ref{tab:compare_ablation_ontonote} report the experimental results on both datasets, where the last column shows the absolute performance drops compared to \GRN{}.

\begin{figure}[!t]
  \centering
  \includegraphics[width=\linewidth]{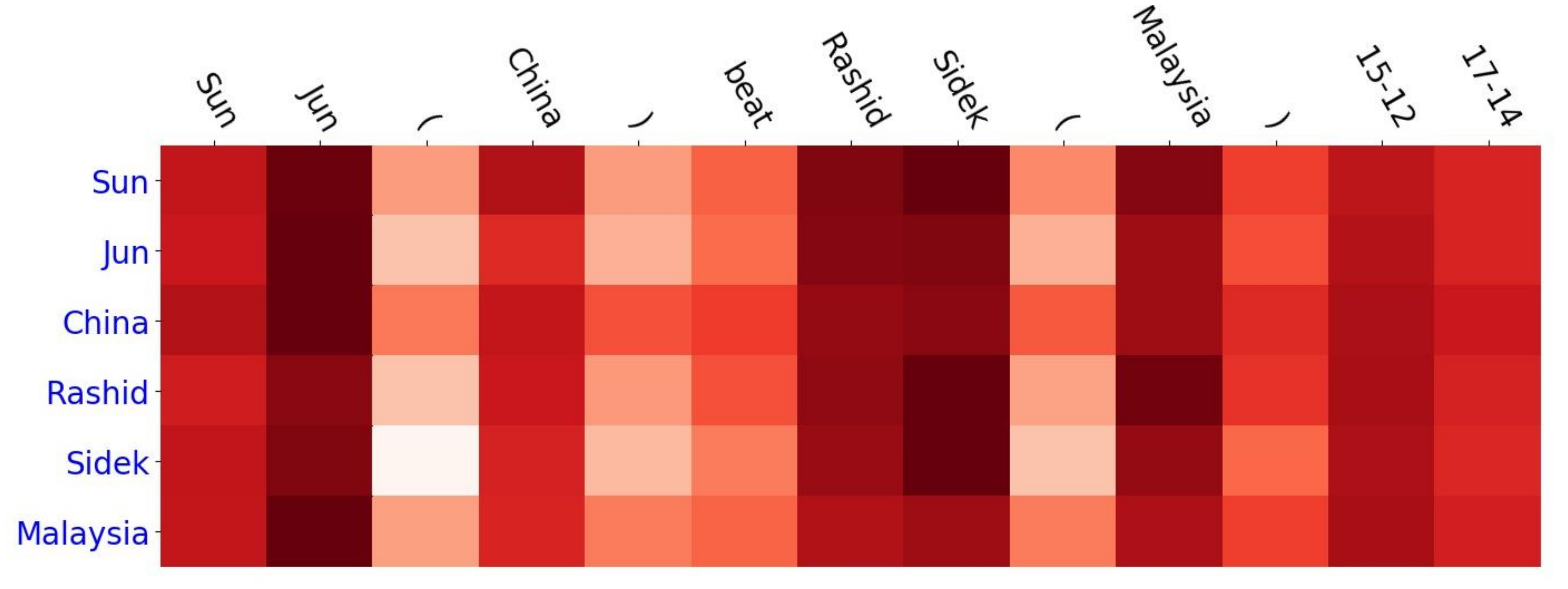}
  \caption{Word relation visualization: the x-axis shows the sentence and the y-axis shows the entity words in it. Regions with deeper color means stronger relations between the corresponding pair of words.}
  \label{fig:visualization}
\end{figure}

As shown in Table~\ref{tab:compare_ablation_conll} and Table~\ref{tab:compare_ablation_ontonote}, when reducing the number of branches in the context layer, \GRN{} w/o context and \GRN{} w/ $\rm branch_3$ drop significantly, which indicates that modelling different scales of local context information plays a crucial role for NER.

Compared with \GRN{} w/o relation, DFN and \GAN{}, the proposed \GRN{} defeats them at a substantial margin in terms of $\rm F_1$ score, which demonstrates that the proposed gated relation layer is beneficial to the performance improvement. The comparison also reveals that the channel-wise gating mechanism in \GRN{} is more powerful than the gated attention approach (\ie{} Eq.~\ref{eq:relation_attention}) and the direct fusion approach (\ie{} Eq.~\ref{eq:relation_sum_up}) under the same experimental settings for NER.

\subsection{Training/Test Time Comparison}

In this section, we further compare the training and test time costs of the proposed \GRN{} with those of CNN-BiLSTM-CRF, which is the most basic LSTM-based NER model achieving high performance. We conduct our experiments on a physical machine with Ubuntu 16.04, 2 Intel Xeon E5-2690 v4 CPUs, and a Tesla P100 GPU. For fair comparison, we keep the representation layer and the CRF layer the same for both models, so that the input and output dimensions for the ``BiLSTM layer'' in CNN-BiLSTM-CRF would be identical to those of the ``context layer + relation layer'' in \GRN{}. We train both models with random initialization for a total of 30 epochs, and after each epoch, we evaluate the learned model on the test set. For both training and test, batch size is set as $10$ as before. And here we use the average training time per epoch and the average test time to calculate speedups.

As shown in Table~\ref{tab:time_complexity}, \GRN{} can obtain a speedup of more than $1.15$ during training and around $1.10$ during test on both datasets. The speedup may seem not so significant, because the time costs reported here also include those consumed by common representation layer, CRF layer, etc. For reference, the fast ID-CNN with a CRF layer (\ie{} ID-CNN-CRF) \cite{strubell2017fast} was reported to have a test-time speedup of $1.28$ over the basic BiLSTM-CRF model on CoNLL-2003. Compared to ID-CNN-CRF, \GRN{} sacrifices some speedup for better performance, and the speedup gap between both is still reasonable. 
We can also see that the speedup on CoNLL-2003 is larger than that on OntoNotes 5.0, which can be attributed to that the average sentence length of CoNLL-2003 ($\sim 14$) is smaller than that of OntoNotes 5.0 ($\sim 18$) and thus the relation layer in \GRN{} would cost less time for the former.
The results above demonstrate that the proposed \GRN{} can generally bring efficiency improvement over LSTM-based methods for NER, via fully exploiting the GPU parallelism.

\begin{table}[t!]
  \centering
    \begin{adjustbox}{width=0.75\columnwidth}
    \scalebox{0.8}{
    \begin{tabular}{|l|c|c|}
      \hline
       & CoNLL-2003 & OntoNotes 5.0  \\ \hline
       Training &1.16x &1.15x \\ \hline
       Test &1.19x &1.08x \\ \hline
    \end{tabular}}
    \end{adjustbox}
    \caption{Training/test speedup of \GRN{} compared with CNN-BiLSTM-CRF. }
    \label{tab:time_complexity}
\end{table}

\subsection{Word Relation Visualization}
Since the proposed \GRN{} aims to boost the NER performance by modelling the relations between words, especially long-term ones, we can visualize the gating output in the relation layer to illustrate the interpretability of \GRN{}. Specifically, we utilize the $L2$ norm of $r_{ij}$ to indicate the extent of relations between the word $s_i$ and the word $s_j$. Then we further normalize the values into $\left[0, 1\right]$ to build a heat map. Figure~\ref{fig:visualization} shows a visualization sample. We can find out that the entity words (y-axis) are more related to other entity words as well, even though they may be ``far away'' from each other in the sentence, like the 1st word ``Sun'' and the 8th word ``Sidek'' in the sample. Note that ``Sun'' and ``Sidek'' are not in an identical receptive field of any CNN used in our experiments, but their strong correlation can still be exploited with the relation layer in \GRN{}. That concretely illustrates that, by introducing the gated relation layer, \GRN{} is able to capture the long-term dependency between words. 


\section{Conclusion}
In this paper, we propose a CNN-based network, \ie{} gated relation network (\GRN{}), for named entity recognition (NER). Unlike the dominant LSTM-based NER models which process a sentence in a sequential manner, \GRN{} can process all the words concurrently with one forward operation and thus can fully exploit the GPU parallelism for potential efficiency improvement. Besides, compared with common CNNs, \GRN{} has a better capacity of capturing long-term context information. Specifically, \GRN{} introduces a gated relation layer to model the relations between any two words, and utilizes gating mechanism to fuse local context features into global ones for all words. Experiments on CoNLL-2003 English NER and Ontonotes 5.0 datasets show that, \GRN{} can achieve state-of-the-art NER performance with or without external knowledge, meaning that using \GRN{}, CNN-based models can compete with LSTM-based models for NER. Experimental results also show that \GRN{} can generally bring efficiency improvement for training and test.

\section{Acknowledgements}
This work is supported by the National Natural Science Foundation of China (No. 61571269).

\bibliographystyle{aaai}
\bibliography{RelationCNN}

\begin{thebibliography}{}

\bibitem[\protect\citeauthoryear{Bikel \bgroup et al\mbox.\egroup
  }{1997}]{bikel1997nymble}
Bikel, D.~M.; Miller, S.; Schwartz, R.; and Weischedel, R.
\newblock 1997.
\newblock Nymble: a high-performance learning name-finder.
\newblock In {\em ANLP},  194--201.

\bibitem[\protect\citeauthoryear{Chen \bgroup et al\mbox.\egroup
  }{2017}]{chen2017sca}
Chen, L.; Zhang, H.; Xiao, J.; Nie, L.; Shao, J.; Liu, W.; and Chua, T.-S.
\newblock 2017.
\newblock Sca-cnn: Spatial and channel-wise attention in convolutional networks
  for image captioning.
\newblock In {\em CVPR},  6298--6306.

\bibitem[\protect\citeauthoryear{Chiu and Nichols}{2016}]{chiu2016named}
Chiu, J., and Nichols, E.
\newblock 2016.
\newblock Named entity recognition with bidirectional lstm-cnns.
\newblock {\em TACL} 4(1):357--370.

\bibitem[\protect\citeauthoryear{Collobert \bgroup et al\mbox.\egroup
  }{2011}]{collobert2011natural}
Collobert, R.; Weston, J.; Bottou, L.; Karlen, M.; Kavukcuoglu, K.; and Kuksa,
  P.
\newblock 2011.
\newblock Natural language processing (almost) from scratch.
\newblock {\em JMLR} 12:2493--2537.

\bibitem[\protect\citeauthoryear{Durrett and Klein}{2014}]{durrett2014joint}
Durrett, G., and Klein, D.
\newblock 2014.
\newblock A joint model for entity analysis: Coreference, typing, and linking.
\newblock {\em TACL} 2(1):477--490.

\bibitem[\protect\citeauthoryear{Gehring \bgroup et al\mbox.\egroup
  }{2017}]{gehring2017convolutional}
Gehring, J.; Auli, M.; Grangier, D.; Yarats, D.; and Dauphin, Y.~N.
\newblock 2017.
\newblock Convolutional sequence to sequence learning.
\newblock In {\em ICML},  1243--1252.

\bibitem[\protect\citeauthoryear{He \bgroup et al\mbox.\egroup
  }{2015}]{he2015delving}
He, K.; Zhang, X.; Ren, S.; and Sun, J.
\newblock 2015.
\newblock Delving deep into rectifiers: Surpassing human-level performance on
  imagenet classification.
\newblock In {\em ICCV},  1026--1034.

\bibitem[\protect\citeauthoryear{Hochreiter and
  Schmidhuber}{1997}]{hochreiter1997long}
Hochreiter, S., and Schmidhuber, J.
\newblock 1997.
\newblock Long short-term memory.
\newblock {\em Neural Computation} 9(8):1735--1780.

\bibitem[\protect\citeauthoryear{Hovy \bgroup et al\mbox.\egroup
  }{2006}]{hovy2006ontonotes}
Hovy, E.; Marcus, M.; Palmer, M.; Ramshaw, L.; and Weischedel, R.
\newblock 2006.
\newblock Ontonotes: the 90\% solution.
\newblock In {\em NAACL-HLT},  57--60.

\bibitem[\protect\citeauthoryear{Huang, Xu, and
  Yu}{2015}]{huang2015bidirectional}
Huang, Z.; Xu, W.; and Yu, K.
\newblock 2015.
\newblock Bidirectional lstm-crf models for sequence tagging.
\newblock {\em arXiv:1508.01991}.

\bibitem[\protect\citeauthoryear{Lample \bgroup et al\mbox.\egroup
  }{2016}]{lample2016neural}
Lample, G.; Ballesteros, M.; Subramanian, S.; Kawakami, K.; and Dyer, C.
\newblock 2016.
\newblock Neural architectures for named entity recognition.
\newblock In {\em NAACL-HLT},  260--270.

\bibitem[\protect\citeauthoryear{Liu, Baldwin, and
  Cohn}{2017}]{liu2017capturing}
Liu, F.; Baldwin, T.; and Cohn, T.
\newblock 2017.
\newblock Capturing long-range contextual dependencies with memory-enhanced
  conditional random fields.
\newblock In {\em IJCNLP},  555--565.

\bibitem[\protect\citeauthoryear{{Liu} \bgroup et al\mbox.\egroup
  }{2018}]{Liu2018Empower}
{Liu}, L.; {Shang}, J.; {Xu}, F.; {Ren}, X.; {Gui}, H.; {Peng}, J.; and {Han},
  J.
\newblock 2018.
\newblock {Empower Sequence Labeling with Task-Aware Neural Language Model}.
\newblock In {\em AAAI}.

\bibitem[\protect\citeauthoryear{Luo \bgroup et al\mbox.\egroup
  }{2015}]{luo2015joint}
Luo, G.; Huang, X.; Lin, C.-Y.; and Nie, Z.
\newblock 2015.
\newblock Joint entity recognition and disambiguation.
\newblock In {\em EMNLP},  879--888.

\bibitem[\protect\citeauthoryear{Ma and Hovy}{2016}]{ma2016CNNBLSTMCRF}
Ma, X., and Hovy, E.
\newblock 2016.
\newblock End-to-end sequence labeling via bi-directional lstm-cnns-crf.
\newblock In {\em ACL},  1064--1074.

\bibitem[\protect\citeauthoryear{McCallum and Li}{2003}]{mccallum2003early}
McCallum, A., and Li, W.
\newblock 2003.
\newblock Early results for named entity recognition with conditional random
  fields, feature induction and web-enhanced lexicons.
\newblock In {\em CoNLL},  188--191.

\bibitem[\protect\citeauthoryear{Passos, Kumar, and
  McCallum}{2014}]{passos2014lexicon}
Passos, A.; Kumar, V.; and McCallum, A.
\newblock 2014.
\newblock Lexicon infused phrase embeddings for named entity resolution.
\newblock In {\em CoNLL},  78--86.

\bibitem[\protect\citeauthoryear{Paszke \bgroup et al\mbox.\egroup
  }{2017}]{paszke2017automatic}
Paszke, A.; Gross, S.; Chintala, S.; Chanan, G.; Yang, E.; DeVito, Z.; Lin, Z.;
  Desmaison, A.; Antiga, L.; and Lerer, A.
\newblock 2017.
\newblock Automatic differentiation in pytorch.
\newblock In {\em NIPS}.

\bibitem[\protect\citeauthoryear{Pennington, Socher, and
  Manning}{2014}]{pennington2014glove}
Pennington, J.; Socher, R.; and Manning, C.
\newblock 2014.
\newblock Glove: Global vectors for word representation.
\newblock In {\em EMNLP},  1532--1543.

\bibitem[\protect\citeauthoryear{Peters \bgroup et al\mbox.\egroup
  }{2017}]{peters2017semi}
Peters, M.; Ammar, W.; Bhagavatula, C.; and Power, R.
\newblock 2017.
\newblock Semi-supervised sequence tagging with bidirectional language models.
\newblock In {\em ACL},  1756--1765.

\bibitem[\protect\citeauthoryear{Peters \bgroup et al\mbox.\egroup
  }{2018}]{peters2018deep}
Peters, M.; Neumann, M.; Iyyer, M.; Gardner, M.; Clark, C.; Lee, K.; and
  Zettlemoyer, L.
\newblock 2018.
\newblock Deep contextualized word representations.
\newblock In {\em NAACL-HLT},  2227--2237.

\bibitem[\protect\citeauthoryear{Pradhan \bgroup et al\mbox.\egroup
  }{2013}]{pradhan2013towards}
Pradhan, S.; Moschitti, A.; Xue, N.; Ng, H.~T.; Bj{\"o}rkelund, A.; Uryupina,
  O.; Zhang, Y.; and Zhong, Z.
\newblock 2013.
\newblock Towards robust linguistic analysis using ontonotes.
\newblock In {\em CoNLL},  143--152.

\bibitem[\protect\citeauthoryear{Rei, Crichton, and
  Pyysalo}{2016}]{rei2016attending}
Rei, M.; Crichton, G.; and Pyysalo, S.
\newblock 2016.
\newblock Attending to characters in neural sequence labeling models.
\newblock In {\em COLING},  309--318.

\bibitem[\protect\citeauthoryear{Rei}{2017}]{rei2017semi}
Rei, M.
\newblock 2017.
\newblock Semi-supervised multitask learning for sequence labeling.
\newblock In {\em ACL},  2121--2130.

\bibitem[\protect\citeauthoryear{Santoro \bgroup et al\mbox.\egroup
  }{2017}]{santoro2017simple}
Santoro, A.; Raposo, D.; Barrett, D.~G.; Malinowski, M.; Pascanu, R.;
  Battaglia, P.; and Lillicrap, T.
\newblock 2017.
\newblock A simple neural network module for relational reasoning.
\newblock In {\em NIPS},  4967--4976.

\bibitem[\protect\citeauthoryear{Shen \bgroup et al\mbox.\egroup
  }{2017}]{shen2017deep}
Shen, Y.; Yun, H.; Lipton, Z.; Kronrod, Y.; and Anandkumar, A.
\newblock 2017.
\newblock Deep active learning for named entity recognition.
\newblock In {\em Workshop on Representation Learning for NLP},  252--256.

\bibitem[\protect\citeauthoryear{Strubell \bgroup et al\mbox.\egroup
  }{2017}]{strubell2017fast}
Strubell, E.; Verga, P.; Belanger, D.; and McCallum, A.
\newblock 2017.
\newblock Fast and accurate entity recognition with iterated dilated
  convolutions.
\newblock In {\em EMNLP},  2670--2680.

\bibitem[\protect\citeauthoryear{Szegedy \bgroup et al\mbox.\egroup
  }{2015}]{szegedy2015going}
Szegedy, C.; Liu, W.; Jia, Y.; Sermanet, P.; Reed, S.; Anguelov, D.; Erhan, D.;
  Vanhoucke, V.; and Rabinovich, A.
\newblock 2015.
\newblock Going deeper with convolutions.
\newblock In {\em CVPR},  1--9.

\bibitem[\protect\citeauthoryear{Tjong Kim~Sang and
  De~Meulder}{2003}]{tjong2003introduction}
Tjong Kim~Sang, E.~F., and De~Meulder, F.
\newblock 2003.
\newblock Introduction to the conll-2003 shared task: Language-independent
  named entity recognition.
\newblock In {\em CoNLL},  142--147.

\bibitem[\protect\citeauthoryear{Tran, MacKinlay, and
  Yepes}{2017}]{tran2017named}
Tran, Q.; MacKinlay, A.; and Yepes, A.~J.
\newblock 2017.
\newblock Named entity recognition with stack residual lstm and trainable bias
  decoding.
\newblock In {\em IJCNLP},  566--575.

\bibitem[\protect\citeauthoryear{Vaswani \bgroup et al\mbox.\egroup
  }{2017}]{vaswani2017attention}
Vaswani, A.; Shazeer, N.; Parmar, N.; Uszkoreit, J.; Jones, L.; Gomez, A.~N.;
  Kaiser, {\L}.; and Polosukhin, I.
\newblock 2017.
\newblock Attention is all you need.
\newblock In {\em NIPS},  5998--6008.

\bibitem[\protect\citeauthoryear{Yang, Liang, and Zhang}{2018}]{yang2018design}
Yang, J.; Liang, S.; and Zhang, Y.
\newblock 2018.
\newblock Design challenges and misconceptions in neural sequence labeling.
\newblock In {\em COLING},  3879--3889.

\bibitem[\protect\citeauthoryear{Yang, Salakhutdinov, and
  Cohen}{2017}]{yang2017transfer}
Yang, Z.; Salakhutdinov, R.; and Cohen, W.~W.
\newblock 2017.
\newblock Transfer learning for sequence tagging with hierarchical recurrent
  networks.
\newblock {\em arXiv:1703.06345}.

\bibitem[\protect\citeauthoryear{Ye and Ling}{2018}]{Ye2018HSCRF}
Ye, Z., and Ling, Z.-H.
\newblock 2018.
\newblock Hybrid semi-markov crf for neural sequence labeling.
\newblock In {\em ACL},  235--240.

\bibitem[\protect\citeauthoryear{Zhuo \bgroup et al\mbox.\egroup
  }{2016}]{zhuo2016segment}
Zhuo, J.; Cao, Y.; Zhu, J.; Zhang, B.; and Nie, Z.
\newblock 2016.
\newblock Segment-level sequence modeling using gated recursive semi-markov
  conditional random fields.
\newblock In {\em ACL},  1413--1423.

\bibitem[\protect\citeauthoryear{Zukov-Gregoric \bgroup et al\mbox.\egroup
  }{2017}]{zukov2017neural}
Zukov-Gregoric, A.; Bachrach, Y.; Minkovsky, P.; Coope, S.; and Maksak, B.
\newblock 2017.
\newblock Neural named entity recognition using a self-attention mechanism.
\newblock In {\em ICTAI},  652--656.

\end{thebibliography}

\end{document}